\documentclass[sigconf]{acmart}

\AtBeginDocument{%
  }

\usepackage[capitalize,noabbrev]{cleveref}
\usepackage{balance}
\usepackage[table]{xcolor}
\definecolor{f2ecde}{HTML}{f2ecde}
\copyrightyear{2026}
\acmYear{2026}
\setcopyright{cc}
\setcctype{by}
\acmConference[MM '26] {Proceedings of the 34th ACM International Conference on Multimedia}{November 10--14, 2026}{Rio de Janeiro, Brazil.}
\acmBooktitle{Proceedings of the 34th ACM International Conference on Multimedia (MM '26), November 10--14, 2026, Rio de Janeiro, Brazil}
\acmISBN{979-8-4007-2213-4/2026/11}
\acmDOI{10.1145/3767308.3836548}
\begin{document}

\title[MultiBinder]{MultiBinder: Preventing Multi-Subject Identity Leakage in Personalized Video Generation}

\author{Wei Li}
\authornote{This work was performed at HiDream.ai.}
\affiliation{%
  \institution{University of Science\\and Technology of China}
  \city{Hefei}
  \country{China}
}
\email{weili2023@mail.ustc.edu.cn}

\author{Yiheng Zhang}
\correspondingauthor
\affiliation{%
  \institution{HiDream.ai Inc.}
  \city{Beijing}
  \country{China}
}
\email{yihengzhang.chn@hidream.ai}

\author{Zhaofan Qiu}
\affiliation{%
  \institution{HiDream.ai Inc.}
  \city{Beijing}
  \country{China}
}
\email{qiuzhaofan@hidream.ai}

\author{Ting Yao}
\affiliation{%
  \institution{HiDream.ai Inc.}
  \city{Beijing}
  \country{China}
}
\email{tiyao@hidream.ai}

\author{Xiaoyan Sun}
\correspondingauthor
\affiliation{%
  \institution{University of Science\\and Technology of China}
  \city{Hefei}
  \country{China}
}
\email{sunxiaoyan@ustc.edu.cn}

\author{Tao Mei}
\affiliation{%
  \institution{HiDream.ai Inc.}
  \city{Beijing}
  \country{China}
}
\email{tmei@hidream.ai}

\renewcommand{\shortauthors}{Wei Li et al.}

\begin{abstract}
Personalized video generation faces a significant scalability challenge: as the number of custom subjects increases, existing methods suffer from severe identity leakage and semantic entanglement, resulting in videos with blended attributes and incoherent narratives. We identify that this stems from a lack of explicit binding between textual entity phrases and their corresponding visual patterns prior to fusion in diffusion models. To address this, we propose MultiBinder, a novel framework built on a ``bind-then-narrate'' paradigm. Our approach introduces a two-stage conditioning process before video diffusion. First, Cross-modal Entity Binding devises an entity-aware Q-Former to extract identity-centric features from each subject’s reference image and tightly align them with its unique entity phrase, yielding disentangled ``identity tokens.'' Second, a Multimodal Large Language Model (MLLM) acts as a ``narrative director'' in the Narrative-Aware Subject Arrangement stage, capturing the action of each subject from the text prompt and synthesizes contextualized subject features that encapsulate ``who is doing what'' and preserve role-specific distinctions. These features are then injected into a video diffusion model via dedicated conditioning pathways, ensuring precise attribute allocation. Extensive experiments on both general and newly curated multi-subject benchmarks show that MultiBinder significantly outperforms existing methods in multi-subject identity fidelity.
\end{abstract}




\maketitle

\section{Introduction}
Personalized video generation aims to produce dynamic video sequences where custom subjects, defined by one or a few reference images, perform actions described in a textual prompt. 
\begin{figure*}[!t]
    \centering
    \includegraphics[width=1.0\linewidth]{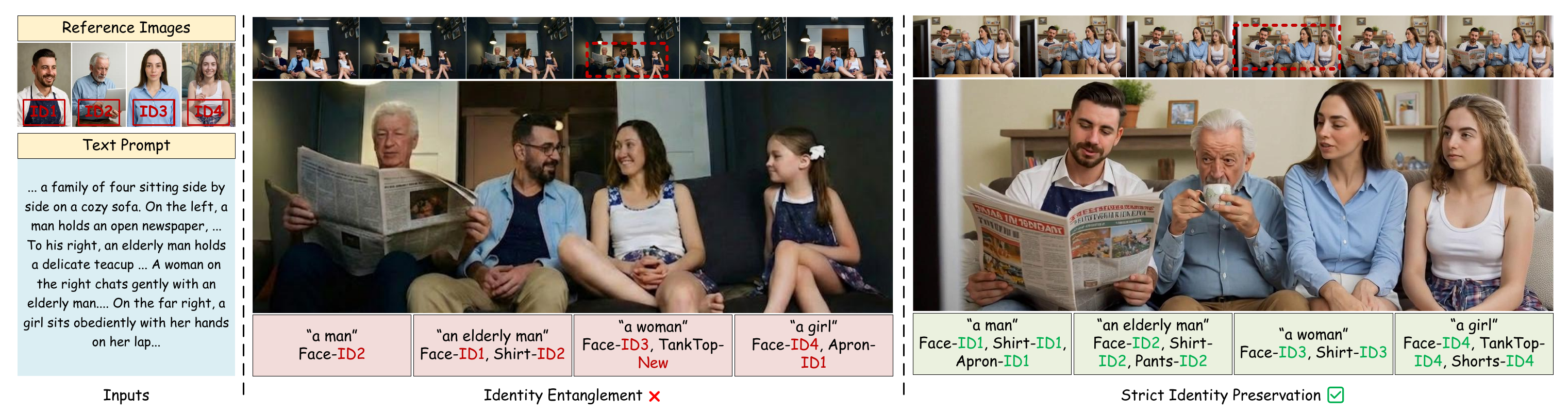}
    \vspace{-0.3in}
    \caption{\small Given a text prompt and four subjects, conventional approaches (top) suffer from ambiguous cross-modal alignment, leading to identity entanglement where visual attributes blend incoherently across subjects (e.g., the “elderly man” incorrectly inherits the face of ID1). MultiBinder (bottom) establishes unambiguous entity bindings, ensuring strict identity preservation.}
    \label{fig:intro}
    \vspace{-0.08in}
\end{figure*}
Successful applications range from creating custom animated avatars and personalized storytelling to dynamic advertising with branded characters. However, the transition from single-subject to multi-subject personalized video generation exposes a critical scalability bottleneck. As shown in Figure~\ref{fig:intro}, when multiple distinct subjects (e.g., the man/elderly man/woman/girl in the figure) are customized within a single generated video, the performance of existing methods degrades significantly. The core failure manifests as severe identity leakage and semantic entanglement: visual attributes from different subjects blend incoherently, creating chimera-like characters with mixed features, while the intended narrative structure collapses.

We identify the root cause of this problem as the ambiguous cross-modal alignment inherent in the standard conditioning paradigm of diffusion-based video models. Typically, reference images and text prompts are fused into a joint conditioning signal without explicit, structured mechanisms to isolate and preserve the correspondence between a textual mention of an entity (e.g., the ``elderly man'' in Figure~\ref{fig:intro}) and its unique visual pattern from the reference set. During the denoising process, especially in attention layers, features from different subjects become entangled, leading to the observed identity blending. The model lacks a definitive, pre-fusion step to bind each subject's semantic concept to its visual identity.

To address this, we propose \textbf{MultiBinder}, a novel framework that introduces an explicit ``bind-then-narrate'' paradigm for scalable multi-subject personalized video generation. Our key insight is to establish unambiguous, disentangled cross-modal bindings before the video diffusion process, and then to orchestrate these bound identities within the narrative context. MultiBinder implements this via a two-stage conditioning mechanism. First, in the \textbf{Cross-modal Entity Binding} stage, we devise an entity-aware Q-Former that establishes the connection between entity phrases and visual references. It extracts identity-centric visual features from the reference image and tightly aligns them with the subject’s unique textual entity phrase, producing a set of disentangled, non-fungible identity tokens. This stage ensures that the core visual signature of each subject is isolated and uniquely linked to its semantic descriptor. Second, these identity tokens, along with the global action prompt, are processed by a Multimodal Large Language Model (MLLM) acting as a ``narrative director'' in the \textbf{Narrative-Aware Subject Arrangement} stage. The MLLM’s role is not to generate visual content but to perform spatio-temporal reasoning: it interprets the dynamic scene, captures the intended action for each subject, and synthesizes contextualized subject features. These features encapsulate ``who is doing what'' by integrating the preserved identity with role-specific spatial and temporal cues, all while maintaining clear distinctions between subjects. Finally, these purified and narratively structured features are injected into a pre-trained video diffusion model via dedicated conditioning pathways, guiding the generation to ensure precise attribute allocation and temporal consistency for each subject throughout the video.

The main contributions of this work are summarized as follows. We propose MultiBinder, an innovative framework that introduces a novel ``bind-then-narrate'' pipeline and enables effective multi-subject preservation in personalized video generation. To validate our framework, we construct a new multi-subject benchmark (MultiBind-Bench) and achieve superior performance over both previous research works and commercial products. Overall, our work directly addresses the critical issue of identity entanglement and attribute leakage in multi-subject personalized video generation, tackling an important yet underexplored challenge in this field.

\section{Related Work}
\textbf{Video generation model.} The field of video generation has witnessed remarkable progress, evolving from early UNet-based diffusion models~\cite{blattmann2023stable,animatediff,xu2025clgc,zhang2024trip,2025motionpro,luo2024freeenhance} to transformer-based models~\cite{dit,sora}. This evolution has been strongly influenced by advances in image generation~\cite{sd3,cai2025hidream,cai2026hidream}. Initial UNet-based video diffusion models often extend image generation models to video domain by integrating temporal modules~\cite{wu2023tune, animatediff}. More recent advancements have introduced diffusion transformer to replace UNet backbone. The rapid evolution has been further accelerated by the emergence of open-source foundational video generation models, such as OpenSora~\cite{opensora}, CogVideoX~\cite{yang2025cogvideox}, HunyuanVideo~\cite{hunyuanvideo} and Wan~\cite{wan2025wan}, which have democratized access to advanced video synthesis capabilities. The powerful base models for text-to-video generation pave the way for personalized video generation.


\noindent\textbf{Personalized video generation.}
Personalized video generation seeks to generate identity-consistent videos based on given reference images~\cite{li2025edit} and text prompt. Existing methods fall into two categories: test-time tuning method~\cite{wu2024motionbooth,customvideo,magic-me,chefer2024still,xu2025smrabooth,xu2026disco}, and tuning-free method~\cite{wei2024dreamvideo2,zhang2025fantasyid,wei2025echovideo,magic-mirror,sugar}. Test-time tuning methods embed specific concept into the output domain of the pre-trained diffusion model via fine-tuning, like DreamBooth~\cite{ruiz2023dreambooth} and Textual Inversion~\cite{gal2022image}. The lengthy fine-tuning for each new concept limits their practicality. Tuning-free methods bypass per-identity fine-tuning. These approaches often focus on more effective identity representation and integration of identity conditions. ConsisID~\cite{consisid} enhances facial consistency via frequency decomposition. DreamVideo~\cite{wei2024dreamvideo} first decouples the learning for subject and motion. Recent works~\cite{li2025create,conceptmaster,chen2025multi,fei2025skyreels,hunyuancustom,liu2025phantom,vace,zhong2025concat,zhang2025region} keep exploring optimal methods for extracting robust visual representations and developing sophisticated conditioning strategies to integrate identity features into the generation process effectively. For example, ConceptMaster~\cite{conceptmaster} employs a Q-Former~\cite{blip2} and CLIP~\cite{radford2021clip} to extract fine-grained visual features, whereas VideoAlchemy~\cite{chen2025multi} adopts DINOv2~\cite{oquab2023dinov2} to capture distinctive subject-specific representations. Both of them lack the high-level integration of text descriptions and reference subjects. BindWeave~\cite{bindweave} and CINEMA \cite{deng2025cinema} both leverage a MLLM to interpret complex context, integrating text and visual features. However, they do not fully exploit the link of textual description and reference entity. Despite these advancements~\cite{polyvivid,magref,lireactid}, a critical challenge remains underexplored: identity blending and attribute leakage problem in multi-subject scenarios. 
\begin{figure*}[!t]
    \centering
    \includegraphics[width=\linewidth]{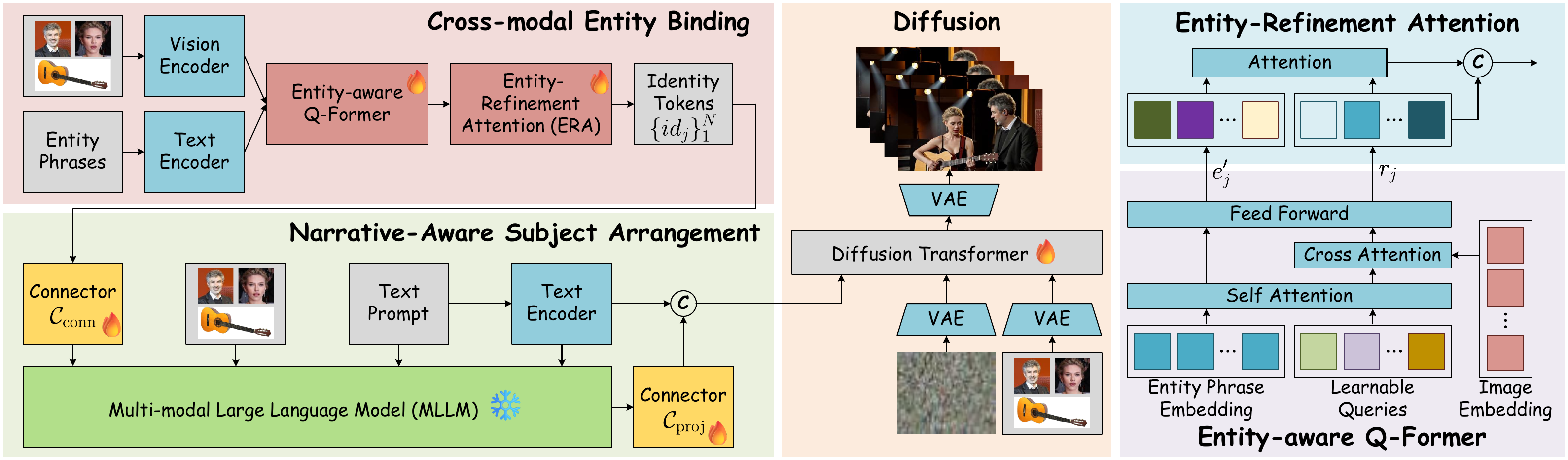}
    \vspace{-0.2in}
    \caption{\small Overview of the MultiBinder framework. The Cross-modal Entity Binding module utilizes an Entity-aware Q-Former and Entity-Refinement Attention to link visual attributes with entity phrases. Subsequently, the MLLM performs Narrative-Aware Subject Arrangement, interpreting the text prompt to contextualize identity tokens for the final Diffusion Transformer.}
    \vspace{-0.05in}
    \label{fig:framework}
\end{figure*}
While current methods show promise in single-subject scenarios, producing videos involving multiple references frequently leads to visual artifacts, confusion of identities, or attribute blending. This persistent limitation highlights a significant gap in current research, impeding the creation of narrative-rich personalized videos that accurately depict interactions among multiple subjects.

\section{Preliminaries}
\noindent\textbf{Diffusion Transformer for Video Generation.}
To leverage the prior knowledge from pre-trained video generation models~\cite{sora,opensora,hunyuanvideo,wan2025wan}, we adopt an advanced video diffusion transformer, i.e., Wan-T2V-1.3B~\cite{wan2025wan}, as the backbone architecture in our approach. Typically, most video DiT models are grounded in flow matching~\cite{flowmatch2022,sd3} theory, which provides a theoretically rigorous framework for learning continuous-time generative processes. 
It aims to learn a vector field that transports a noise sample $x_0$ to a clean data sample $x_1$ with straight path $x_t = t x_1 + (1-t) x_0$ at timestamp $t$ based on Rectified Flow~\cite{sd3}, where $x_0 \sim \mathcal{N}(0, 1)$ is a normal gaussian noise. 
Then, the ground-truth velocity vector is calculated as:
\begin{equation}
v_t = \frac{dx_t}{dt} = x_1 - x_0 .
\label{eq:vt}
\end{equation}
The video DiT model is learned to estimate the velocity vector via:
\begin{equation}
u_\theta(t) = u(x_t, c, t; \theta) ,
\label{eq:u}
\end{equation}
where $\theta$ represents the model parameters, and $c$ is the set of input conditions. Therefore, the flow matching training objective $\mathcal{L}$ is defined as the mean squared error (MSE) between the model's output and the ground-truth velocity $v_t$:
\begin{equation}
\small
\mathcal{L} = \mathbb{E}_{x_0, x_1, c, t} \bigl\| u(x_t, c, t; \theta) - v_t \bigr\|^2 .
\label{eq:loss}
\end{equation}

\noindent\textbf{Multimodal Large Language Models.}
Multimodal Large Language Models (MLLMs)~\cite{liu2023visual,qwen2.5,blip2} aim to equip Large Language Models (LLMs)~\cite{gpt4} with cross-modal understanding and generation capabilities across various modalities~\cite{liu2025javisdit}. A typical MLLM architecture (e.g., LLaVA~\cite{liu2023visual}) primarily consists of three core components: a pre-trained modality-specific encoder, such as CLIP~\cite{radford2021clip} for visual input and Whisper~\cite{whisper} for audio input, a foundation LLM~\cite{qwen2.5,qwen3} that serves as the central reasoning module, and a multimodal projector that bridges the representational gap between modality-specific encoders and the LLM's token embedding space.

\section{MultiBinder}
We propose MultiBinder, a novel entity-aware personalized video generation framework designed to tackle the persistent challenge of identity blending and attribute leakage in multi-subject scenarios. 
Operating on a ``bind-then-narrate'' paradigm, MultiBinder unfolds two stages, as illustrated in Figure~\ref{fig:framework}. 
First, in \textbf{Cross-modal Entity Binding}, given reference images and entity phrases, we employ an entity-aware Q-Former to disentangle core visual attributes and bind them to their corresponding textual entities, yielding distinct identity tokens. Second, during \textbf{Narrative-Aware Subject Arrangement}, we leverage a Multimodal Large Language Model (MLLM) as a reasoning engine. It interprets the text prompt to contextualize these identity tokens, ensuring precise role allocation and spatio-temporal consistency.
Finally, these purified and narratively structured features are injected into a Diffusion Transformer (DiT) to guide the generation during the denoising process.

In the following subsections, we detail the core components of our framework. Section \ref{sec:binding} elaborates on the Cross-modal Entity Binding mechanism for extracting distinct identity tokens. Section \ref{sec:narrative} describes the Narrative-Aware Subject Arrangement driven by the MLLM. Section \ref{sec:training} introduces detailed training strategy in Multibinder, followed by our proposed MultiBind-Bench for evaluation in Section \ref{sec:benchmark}.

\subsection{Cross-modal Entity Binding}\label{sec:binding}
To enable precise multi-subject identity preservation for video generation, the Diffusion Transformer (DiT) must be conditioned on persistent visual representations that are uniquely and unambiguously linked to their specific textual entities (i.e., entity phrases). Conventional visual encoders like CLIP~\cite{radford2021clip} and DINO~\cite{oquab2023dinov2} typically yield either global semantic embeddings or raw patch-level features, both of which are suboptimal for this task. Global embeddings often lack fine-grained identity details and are susceptible to background noise, failing to provide sufficient information for high-fidelity reconstruction. Conversely, while patch-level features preserve local details, the specific attributes of an entity are uncontrollably scattered across the feature sequence. During the generation process, the DiT struggles to align these dispersed visual cues with specific concepts; consequently, without explicit binding, attributes from one entity may inadvertently associate with the entity phrase of another subject in the text prompt. This misalignment directly leads to identity leakage and suboptimal generation results. To address this, we introduce the Cross-modal Entity Binding module, which leverages an entity-aware Q-Former augmented by Entity-Refinement Attention (ERA) to lock visual attributes to their corresponding entity phrases.

Formally, for a personalized video generation task involving $N$ reference images $\mathcal{I}=\{ I_i \}_{i=1}^{N}$, we derive a set of $N$ entity phrases $\{ E_j \}_{j=1}^{N}$ from the text prompt $text$. We first employ a frozen CLIP image encoder $\mathcal{E}_{\text{CLIP}}$ to extract dense visual features $f_i = \mathcal{E}_{\text{CLIP}}(I_i)$. Simultaneously, a T5 \cite{raffel2020t5} text encoder $\mathcal{E}_{\text{T5}}$ encodes the entity phrases to obtain text embeddings $e_j = \mathcal{E}_{\text{T5}}(E_j)$.
To extract entity-relevant visual features specific to an entity phrase $E_j$, we first concatenate the visual features from all reference images into a unified sequence $F = [f_1, \dots, f_N]$. We tailor the Q-Former to accept the text embedding $e_j$ alongside the learnable queries. Within each transformer layer, the text and query sequences are concatenated to perform self-attention, injecting semantic conditions into the queries. Subsequently, the sequences are split: the queries perform cross-attention over the global visual sequence $F$ to aggregate matched visual patterns, while the text sequence bypasses this step. Both streams are then processed by the FFN. This process initiates the cross-modal binding, extracting both a distinct visual signature $r_j$ that isolates the target subject from the collective visual context, together with the contextualized text embedding $e'_j$. 

To further verify and refine the correspondence between $r_j$ and $e'_j$, we introduce the Entity-Refinement Attention (ERA) module. This module treats the extracted visual signature $r_j$ as the query, and the contextualized text embedding $e'_j$ as both the key and value. By forcing the visual features to attend once more to their rigid semantic definition, the ERA acts as a verification mechanism: it filters out visual cues that do not align with the entity phrase, thereby solidifying the binding. The output is concatenated with the original visual signature $r_j$ to yield the final identity token $id_j$:
\begin{equation}
\small
\begin{aligned}
    & Q = W_{Q} r_j, \quad K = W_K e'_j, \quad V = W_V e'_j, \\
    & id_j = \text{Concat}\left( r_j, \; \text{Softmax}\left(\frac{QK^{T}}{\sqrt{d}}\right) V \right)
\end{aligned}
\end{equation}
where $W_Q, W_K, W_V$ are learnable projection matrices. This process ensures that $id_j$ is not only visually rich but also semantically pure.

\subsection{Narrative-Aware Subject Arrangement}\label{sec:narrative}
While the Entity Binding module effectively isolates the visual appearance of each subject, these identity tokens $\{ id_j \}$ remain static and detached from the temporal dynamics of the video. To bridge this gap, we leverage a frozen Multimodal Large Language Model (MLLM) not merely as an encoder, but as a narrative reasoning engine. By tapping into its extensive world knowledge, the MLLM performs spatio-temporal reasoning: it interprets the dynamic scene defined in the prompt, captures the intended action for each subject, and synthesizes contextualized subject features that serve as a high-level condition for the DiT.

To ground the MLLM's reasoning in the specific visual identities of the users, we first construct a sequence of visual priors. The extracted identity tokens $\{ id_j \}_{j=1}^{N}$ are projected into the MLLM's input space via a trainable connector $\mathcal{C}_{\text{conn}}$. We concatenate these projected tokens to form the identity prefix embedding $E_\text{id}$:
\begin{equation}
\small
    E_\text{id} = [\mathcal{C}_{\text{conn}}(id_1), \mathcal{C}_{\text{conn}}(id_2), \dots, \mathcal{C}_{\text{conn}}(id_N)] ~.
\end{equation}
This prefix acts as a ``character sheet,'' informing the MLLM of the specific appearances of the subjects involved in the narrative.

Next, to incorporate reference images into the reasoning process, we formulate a multimodal instruction sequence $\mathcal{S}$. We append visual placeholders to the text prompt $text$, where each placeholder $\langle\text{img}\rangle_j$ acts as a dedicated slot for the $j$-th reference image $I_j$:
\begin{align}
\small
\mathcal{S} \;&=\; \big[\, text,\; \langle\text{img}\rangle_1, \dots, \langle\text{img}\rangle_N \,\big]~.
\end{align}
During the embedding phase, these placeholders are replaced by the visual embeddings of the corresponding images encoded by the MLLM's vision encoder. By prepending the identity prefix $E_\text{id}$ to this sequence, we enable the MLLM to jointly reason over the text prompt, reference images, and identity priors, aligning the static identities with the dynamic actions described in $text$. The resulting hidden states $H_{\text{mllm}}$ are no longer generic text embeddings, but contextualized subject features that encode the inferred interactions, spatial relationships, and temporal dynamics of the entities.

To transfer this high-level reasoning to the video generation process, $H_{\text{mllm}}$ is projected into the DiT's feature space via a second connector $\mathcal{C}_{\text{proj}}$, yielding the semantic condition $c_{\text{mllm}} = \mathcal{C}_{\text{proj}}(H_{\text{mllm}})$.
Finally, to ensure that the DiT also adheres to precise lexical details and fine-grained syntax, we complement the MLLM features with standard text embeddings from a T5 encoder, $c_{\text{T5}} = \mathcal{E}_{\text{T5}}(text)$. The final conditioning signal $c_{\text{joint}}$ is obtained by concatenating these complementary streams: $c_{\text{joint}} = \text{Concat}(c_{\text{mllm}}, c_{\text{T5}})$.

\subsection{Training of MultiBinder}
\label{sec:training}
We integrate the proposed modules into a Diffusion Transformer (DiT) backbone, optimized using the Rectified Flow formulation \cite{sd3}.
To enable precise narrative control and high-fidelity identity preservation, the DiT is conditioned on a hybrid signal set consisting of: (1) the joint semantic condition $c_{\text{joint}}$, which encapsulates the MLLM-reasoned narrative dynamics and textual instructions; and (2) the visual reference condition $c_{\text{vae}} = \{ \mathcal{E}_{\text{VAE}}(I_j) \}_{j=1}^{N}$, which provides low-level features of the reference entities.
The MultiBinder $u_{\Theta}$ is trained to predict the velocity field $v_t$ that transports the noise distribution to the data distribution. The optimization objective is defined as the mean squared error between the model prediction and the target velocity:
\begin{equation}
\small
\mathcal{L}_{\text{mse}} = \mathbb{E}_{t, z_t, c_{\text{joint}}, c_{\text{vae}}} \left[ \left\| u_{\Theta}(z_t, t, c_{\text{joint}}, c_{\text{vae}}) - v_t \right\|_2^2 \right].
\end{equation}

During training, we adopt a parameter-efficient strategy: the pre-trained image encoders, text encoders, VAE, and the MLLM/Q-Former backbone remain frozen to preserve their generalization priors. Only the learnable queries of Q-Former, the Entity-Refinement Attention, the lightweight connectors ($\mathcal{C}_{\text{conn}}, \mathcal{C}_{\text{proj}}$), and the diffusion transformer backbone are updated.

\subsection{MultiBind-Bench}
\label{sec:benchmark}

Identity blending is a prevalent failure mode in multi-subject video generation, where visual attributes of one subject inadvertently bleed into another. Existing metrics, such as face similarity or NexusScore~\cite{opens2v}, typically operate on pre-matched pairs assuming a clear one-to-one correspondence. Consequently, these matching-based approaches struggle to sensitively capture the occurrence and severity of identity entanglement. To address this, we introduce MultiBind-Bench, a rigorous benchmark designed to explicitly quantify the model's capability to disentangle and correctly assign visual identities.

\noindent \textbf{Evaluation Dataset.}
We curate a dataset comprising 100 multi-subject entries, involving a total of 120 reference images. Each entry consists of 2 to 4 distinct reference subjects and a text prompt. To ensure the consistency and reproducibility of the evaluation, we explicitly provide the corresponding entity phrase for each subject, which serves as ground-truth metadata for detection-based metrics. Note that for practical in-the-wild deployment, these phrases can be automatically extracted from the text prompt using MLLMs such as Qwen2.5-VL. Our evaluation dataset features subject pairs spanning a wide spectrum of visual similarity, ranging from clearly distinguishable identities to highly similar ones. This graduated design enables fine-grained evaluation of the model's capacity to disentangle identity-specific features under varying degrees of ambiguity. In contrast to existing benchmarks such as OpenS2V-Eval, which predominantly assess the holistic quality and general performance of personalized video generation, MultiBind-Bench fills a critical gap by explicitly targeting the largely overlooked challenge of multi-identity disentanglement evaluation.

\noindent \textbf{Identity Purity Score ($S_{\text{IPS}}$).}
We propose Identity Purity Score ($S_{\text{IPS}}$), an alignment-free metric grounded in VQA and entropy to assess global identity affinity without presuming specific matches. To facilitate this evaluation, our dataset explicitly includes discriminative attribute profiles for each reference subject. Specifically, we use Qwen2.5-VL to pre-generate a unified multiple-choice query for each key attribute (hair, face, and clothing) by aggregating the features of all $N$ reference images in the entry. For instance, given the hair attributes of $N$ references, we construct a single question where each option corresponds to the specific description of one reference (e.g., ``Option A: blonde wavy hair (Ref 1); Option B: black straight hair (Ref 2)...'').

The calculation is performed on a frame-by-frame basis. We uniformly sample $T=32$ frames from the generated video and process each frame as follows:

\textit{(1) Visual Verification:} For the $t$-th frame, we employ GroundingDINO~\cite{groundingdino} to localize subjects using the ground-truth entity phrases. The cropped subject regions are then interrogated by Qwen2.5-VL using the pre-defined attribute questions.

\textit{(2) Affinity Matrix Construction:} For each attribute, the VQA model outputs a probability distribution over the $N$ options. We directly interpret the probability assigned to the $n$-th option as the attribute-specific similarity score quantifying the visual correspondence between the generated subject and the $n$-th reference image. We then aggregate these scores across the three attributes via their geometric mean to form the frame-specific affinity matrix $S^{(t)} \in \mathbb{R}^{M \times N}$ (where $M$ denotes the number of detected subjects). The entry $S^{(t)}_{m,n}$ represents the holistic similarity between the $m$-th generated subject and the $n$-th reference:
\begin{equation}
\small
S^{(t)}_{m,n} = \sqrt[3]{s_{m,n}^{\text{hair}} \cdot s_{m,n}^{\text{face}} \cdot s_{m,n}^{\text{cloth}}}~.
\end{equation}

\textit{(3) Entropy-based Scoring:} We quantify the identity purity for frame $t$ by computing the mean normalized entropies of $S^{(t)}$ along rows (measuring subject specificity) and columns (measuring reference uniqueness). The score for the current frame is defined as:
\begin{equation}
\small
    \mathcal{S}_t = 1 - (\bar{H}_{\text{row}}^{(t)} + \bar{H}_{\text{col}}^{(t)})/2 ~.
\end{equation}
The final score is derived by averaging the scores across all sampled frames: $S_{\text{IPS}} = \frac{1}{T} \sum_{t=1}^{T} \mathcal{S}_t$.

\noindent \textbf{Visual Reasoning Score ($S_{\text{VRS}}$).}
To supplement the entropy-based metric, we introduce a high-level evaluator designed to scrutinize subtle identity entanglements that require advanced cognitive judgment. We term this metric the Visual Reasoning Score ($S_{\text{VRS}}$). In our implementation, we leverage GPT-4o to perform self-reflective probing. Specifically, for the generated video, we uniformly sample $T=32$ frames, denoted as $\{ V_t \}_{t=1}^{T}$. The model $M_{\text{Reason}}$ evaluates each frame $V_t$ through a Chain-of-Thought (CoT) trajectory. It first formulates recursive probing questions regarding potential attribute leakage, followed by a fine-grained comparative analysis between the reference images $\{ I_j \}$ and the generated frame $V_t$. The judgment is quantified on a 5-point Likert scale (1: worst, 5: best). To ensure stability, the evaluation is repeated twice with independent CoT paths ($k=1, 2$). The final score is normalized to $[0, 1]$:
\begin{equation}
\small
    S_{\text{VRS}} = \frac{1}{4} \left( \left[ \frac{1}{2T} \sum_{k=1}^{2} \sum_{t=1}^{T} M_{\text{Reason}}(V_t) \right] - 1 \right) ~.
\end{equation}

\section{Experiments}
\subsection{Experimental Settings}
\noindent \textbf{Training data.}
We construct a dedicated training dataset through a comprehensive pipeline of web video collection and multi-stage annotation. The procedure includes: (1) downloading candidate videos, (2) filtering based on video content, (3) annotating each video with object detection, segmentation, and descriptive captions, and (4) a final round of annotation verification. The detailed dataset construction protocol is provided in the Appendix. Through this process, we obtain 500k videos, each accompanied by multi-object annotations and a textual description of the video content.

\begin{table*}[!t]
\caption{\small Performance comparisons with state-of-the-art approaches on the OpenS2V-Eval open domain in terms of Aesthetics, Motion Smoothness, Motion Amplitude, Face Similarity, GmeScore, NexusScore, and NaturalScore. The best-performing result is highlighted in \textbf{bold}, and the second-best is \underline{underlined}.}
\label{tab:main_exp}
\vspace{-0.05in}
\centering
\small
\begin{tabular}{l|ccccccc|c}
\toprule
Method & Aes.$\uparrow$ & M. Smo.$\uparrow$ & M. Amp.$\uparrow$ & FaceSim$\uparrow$ & Gme.$\uparrow$ & Nexus.$\uparrow$ & Natural.$\uparrow$ & Total.$\uparrow$ \\
\midrule
VACE-P1.3B     & 47.34\% & 96.8\% & 12.03\% & 16.59\% & \textbf{71.38\%} & 40.19\% & 64.31\% & 48.98\%\\    
VACE-1.3B     & \textbf{48.24\%} & \textbf{97.2\%} & \underline{18.83\%} & 20.57\% & \underline{71.26\%} & 37.91\% & \textbf{65.46\%} & 49.89\% \\
Phantom-1.3B   & 46.67\% & 93.3\% & 14.29\% & \textbf{48.56\%} & 69.43\% & 42.48\% & 62.5\% & \underline{54.89\%}\\
\rowcolor[gray]{0.95} \textbf{MultiBinder} & \underline{48.13\%} & \underline{97.06\%} & \textbf{32.16\%} & \underline{43.45\%} & 69.89\% & \textbf{43.17\%} & \underline{65.18\%} & \textbf{55.52\%} \\
\bottomrule
\end{tabular}
\end{table*}

\noindent \textbf{Evaluation benchmarks and metrics.}
We evaluate our method from two complementary perspectives. First, we test its generalizability on common personalized video generation using the \textbf{OpenS2V-Eval} benchmark \cite{opens2v}. While this benchmark does not focus on the multi-subject entanglement problem addressed in our work, it verifies the framework’s compatibility and performance in general scenarios. OpenS2V-Eval comprises 180 subject–text pairs across seven categories. Following its official protocol, we report seven metrics spanning five dimensions: Aesthetics (visual quality), Motion Smoothness and Motion Amplitude (motion dynamics), Face Similarity and NexusScore (identity consistency), GmeScore (text–video alignment), and NaturalScore (naturalness).

Second, we introduce \textbf{MultiBind-Bench}, a new benchmark with 100 subject–text pairs specifically designed to evaluate multi-subject settings. It directly measures whether visual attributes are correctly assigned to each subject and whether identity leakage occurs. We report two key metrics: Identity Purity Score $S_{\text{IPS}}$ and Visual Reasoning Score $S_{\text{VRS}}$.

\noindent \textbf{Implementation details.}
Our framework builds on Wan2.1-T2V-1.3B \cite{wan2025wan} as the video diffusion backbone. For the MLLM, we use Qwen2.5-VL-3B-Instruct and keep its parameters frozen throughout training. Each adapter consists of a two-layer MLP with GELU activation. We optimize the DiT and adapters with AdamW in a two-stage training scheme: 1) Only the adapters are trained for 4,000 steps (learning rate $1 \times 10^{-5}$) to achieve initial alignment. 2) The DiT and adapters are jointly fine-tuned for 7,000 steps. Both stages include a 100-step linear warm-up. Training is conducted on 16 NVIDIA A100 GPUs with a global batch size of 32. During inference, we use 50 denoising steps and a CFG scale of 5.0.

\subsection{Quantitative Comparisons}
Table \ref{tab:main_exp} presents comparative results on the OpenS2V-Eval benchmark. We evaluate against state-of-the-art methods built on the same Wan2.1-T2V-1.3B backbone: VACE-P1.3B, VACE-1.3B \cite{vace}, and Phantom-1.3B \cite{liu2025phantom}. Despite differences in training data, the comparison remains fair given the shared model architecture and comparable data scales. Our MultiBinder achieves the highest overall score of 55.52\%, significantly surpassing the previous best model (Phantom-1.3B, 54.89\%). These results underscore that our ``bind-then-narrate'' paradigm consistently elevates generation quality across both single- and multi-subject scenarios. By explicitly aligning visual patterns with entity phrases and employing the MLLM for narrative arrangement, our method enhances identity consistency, as reflected in the higher NexusScore. Notably, we observe a clear gain in Motion Amplitude, indicating a reduction in the ``copy-paste'' tendency. The model generates more dynamic and plausible motions beyond replicating the reference pose, benefiting from a deeper semantic understanding of each subject. 

\begin{table}[!t]
\centering
\small
\caption{\small Performance comparisons with state-of-the-art approaches including open-source baselines with varying parameter scales and the commercial system Kling-1.6 on MultiBind-Bench. Results are reported in terms of Identity Purity Score ($S_{\text{IPS}}$) and Visual Reasoning Score ($S_{\text{VRS}}$). The best-performing result is highlighted in \textbf{bold}, and the second-best is \underline{underlined}.}
\label{tab:main_exp2}
\vspace{-0.05in}
\begin{tabular}{l|ccccccc|c}
\toprule
Method & $S_{\text{IPS}}$$\uparrow$ & $S_{\text{VRS}}$$\uparrow$ \\
\midrule
VACE-P1.3B     & 46.37\% & 61.20\% \\    
VACE-1.3B     & 54.64\% & 63.09\% \\ 
Phantom-1.3B   & 55.03\% & 72.15\% \\ 
VACE-14B & 62.38\% & 77.10\% \\ 
Phantom-14B & 65.46\% & 76.44\% \\ 
Kling-1.6 & 69.51\% & 82.89\% \\ 
\rowcolor[gray]{0.95} \textbf{MultiBinder} & \textbf{73.98\%} & \textbf{84.56\%} & \\
\bottomrule
\end{tabular}
\vspace{-0.19in}
\end{table}

Table \ref{tab:main_exp2} reports results on MultiBind-Bench, which is specifically designed to evaluate challenging multi-subject scenarios prone to identity blending. Our method achieves the highest scores on both metrics: 73.98\% in Identity Purity Score and 84.56\% in Visual Reasoning Score. These results not only exceed other Wan2.1-T2V-1.3B–based approaches but also outperform methods built on larger (14B) base models and the commercial system Kling-1.6. Compared with the best competitor (Kling-1.6), our improvements are +4.47\% and +1.67\%, respectively. 
This significant margin confirms that identity leakage remains a critical bottleneck in existing models and validates the core contribution of our framework in effectively addressing this issue. 

\subsection{Qualitative Comparisons}

\begin{figure*}[!t]
    \centering
    \includegraphics[width=0.97\linewidth]{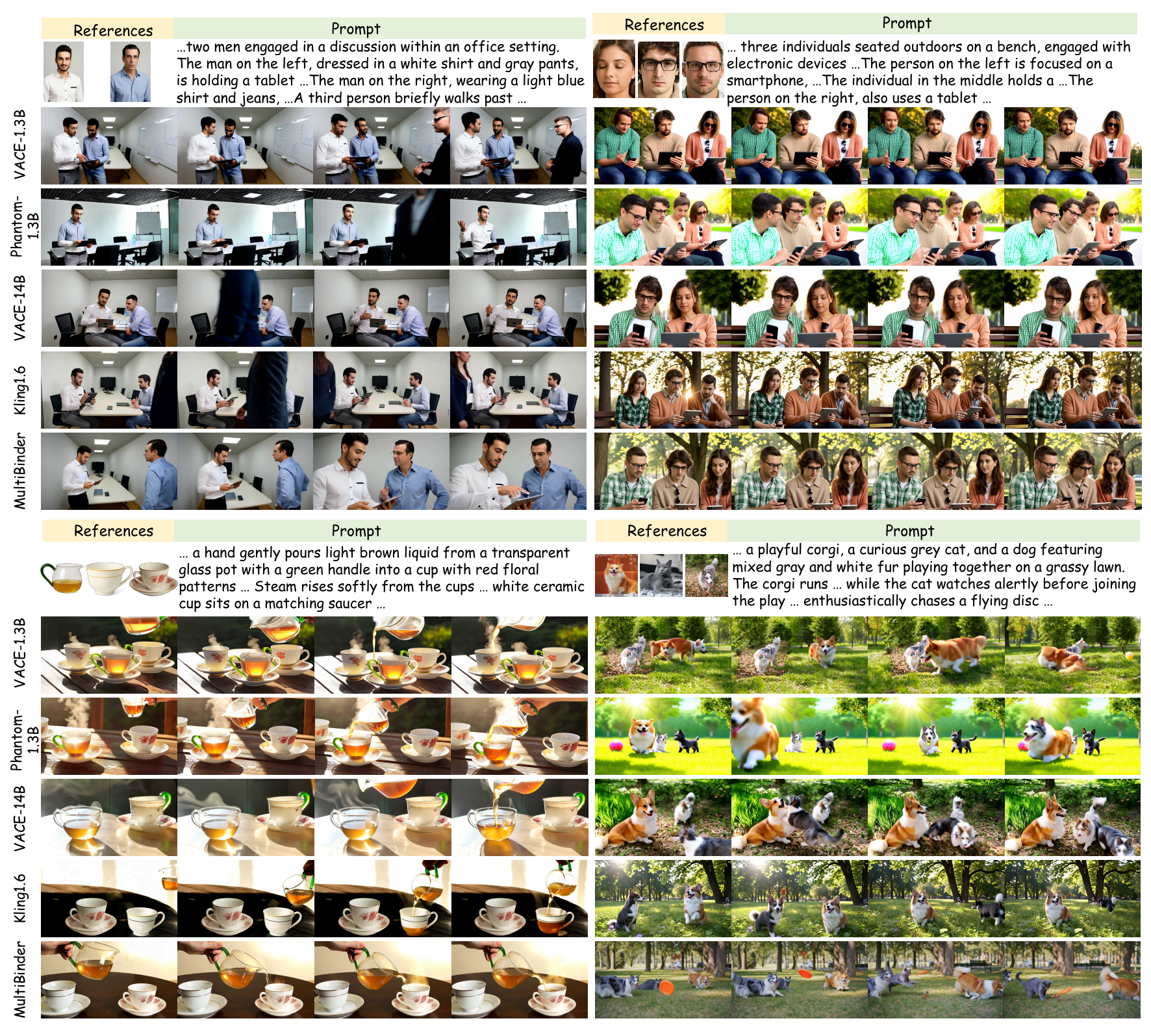}
    \vspace{-0.15in}
    \caption{\small Examples of videos generated by different approaches with multi-subject reference images.}
    \label{fig:quali1}
\vspace{-0.15in}
\end{figure*}

Figure \ref{fig:quali1} provides a qualitative comparison of video results generated by different methods, including our MultiBinder, other 1.3B-based approaches, a 14B-based method (VACE-14B), and a commercial model (Kling-1.6). We present four challenging cases: a two-person scene, a three-person scene, a multi-object scene (teapot and cup), and a multi-animal scene (cat and dog).

In the two-person scenario, all baseline methods exhibit identity preservation issues. For instance, VACE-1.3B, VACE-14B, and Kling-1.6 bleed the visual attributes of the man in white into the feature of the man in blue, while Phantom-1.3B blends the features of both subjects into a single person. This failure mode becomes more severe in the three-person case: Phantom-1.3B and VACE-14B produce an incorrect number of characters, and although VACE-1.3B and Kling-1.6 generates three figures, their appearance no longer matches the reference images due to feature blending. In contrast, our method successfully maintains clear visual identity separation and correct character arrangement in both cases, which we attribute the improvements to our cross-modal entity binding and narrative-aware subject arrangement.

The last two cases extend evaluation to non-human subjects, revealing that identity leakage is a pervasive phenomenon not exclusive to human identities. When visually similar and semantically related pairs—such as teapots and cups or cats and dogs—appear together, baseline methods frequently produce visually blended objects with entangled attributes. Our approach again resolves this feature entanglement effectively in both cases, demonstrating generalizability across diverse subject types beyond human subjects.

\begin{figure}[!t]
    \centering
    \includegraphics[width=0.5\textwidth]{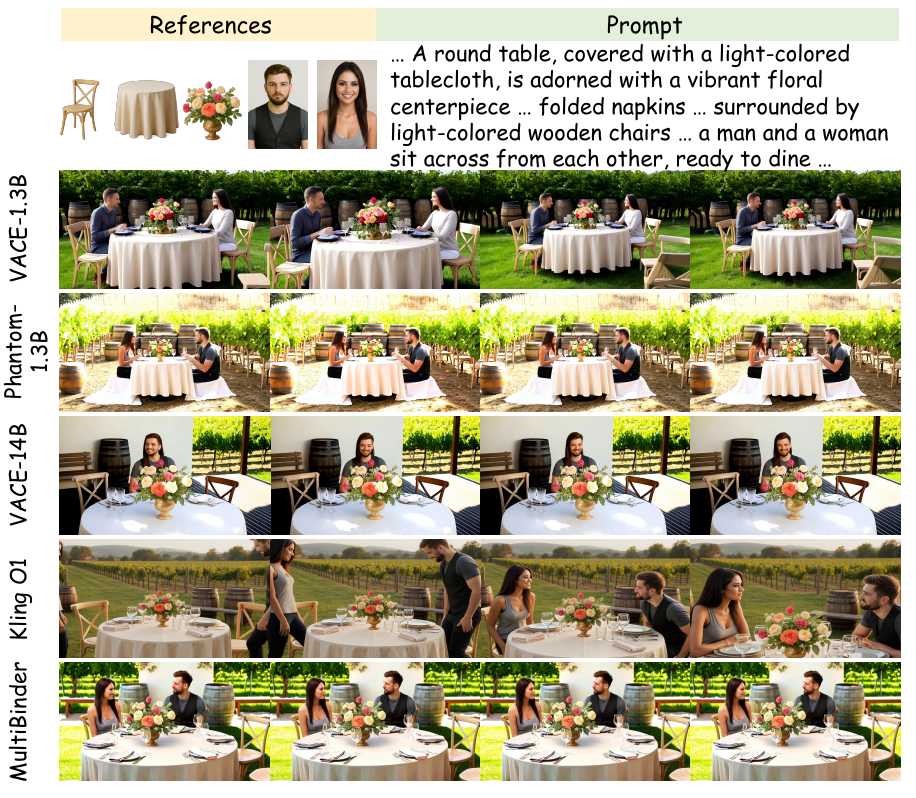}
    \vspace{-0.25in}
    \caption{\small Examples of videos generated by different approaches on scenarios with an increased number of subjects.}
    \label{fig:quali3}
\end{figure}

To further evaluate scalability with respect to subject count, we construct an even more challenging setting involving five distinct subjects: two persons, a table, a chair, and a flower on the table, as shown in Figure \ref{fig:quali3}. This compositionally complex, interactive scene rigorously tests a model's capacity for precise multi-subject assignment under high semantic density. Under this demanding condition, baseline methods exhibit clear and varied failures: VACE-14B merges the appearances of the two persons into one; VACE-1.3B incorrectly alters the characters’ clothing; Phantom-1.3B conflates the visual features of the table and chair. While the commercial model KlingO1 (Kling-1.6 does not support $>4$ subjects input) performs best among baselines—correctly rendering the visual attributes of each subject—it still fails to follow the prompt precisely, generating two chairs instead of the specified three. 
In contrast, MultiBinder exhibits superior generalization, precisely preserving all five identities with accurate attributes and counts. This disparity underscores the indispensability of our MLLM-driven narrative-aware subject arrangement, which leverages MLLM to perform high-level spatio-temporal reasoning and thus effectively orchestrate textual descriptions with reference subject features.

\subsection{Ablation Studies}
We conduct ablation studies on 50\% subset of our training data to validate the contribution of each design choice in MultiBinder. Table~\ref{tab:ablation} reports the performance impact of different components in both stages, evaluated using the total score on OpenS2V-Eval and the Identity Purity Score (IPS) and Visual Reasoning Score (VRS) on MultiBind-Bench.

\noindent\textbf{Cross-Modal Entity Binding.}
We first study the components within the Cross-Modal Entity Binding stage. Replacing the binding module with a vanilla Q-Former yields only a marginal gain of 0.74\% in total score over the baseline without Q-Former. In contrast, our Entity-Aware Q-Former significantly improves the total score by 3.22\%, confirming that explicit entity-level guidance is critical for establishing accurate cross-modal correspondences. While the addition of Entity-Refinement Attention (ERA) does not yield appreciable gains in the total score on general benchmarks, it delivers a substantial +3.15\% improvement in IPS. This result underscores the pivotal role of ERA in mitigating feature interference and refining identity isolation—precisely addressing the fundamental challenge of multi-identity disentanglement.

\noindent\textbf{Narrative-Aware Subject Arrangement.} To isolate the contribution of our second stage, we compare a direct condition baseline without employing a MLLM against our proposed Narrative-Aware Subject Arrangement strategy. Our specialized arrangement design achieves superior performance across all metrics, with improvements of +2.10\% in total score, +4.91\% in IPS, and +3.71\% in VRS. The particularly pronounced gains in IPS and VRS indicate that the benefit is concentrated precisely in identity disentanglement and compositional understanding—the two aspects where naive direct condition strategy is most deficient. These results demonstrate that the observed performance gains do not merely stem from incorporating an MLLM, but rather from our holistic narrative orchestration design, which explicitly harnesses the MLLM for spatio-temporal reasoning and precise per-subject placement under complex compositional constraints.

\begin{table}[!t]
\centering   
\caption{\small Ablation studies on the impact of each design in MultiBinder. Training in this table uses 50\% data.}
\vspace{-0.08in}
\small
\setlength{\tabcolsep}{1.0mm}
\begin{tabular}{l c c c}
\toprule
Method & TotalScore$\uparrow$ & $S_{\text{IPS}}$$\uparrow$ & $S_{\text{VRS}}$$\uparrow$  \\
\midrule
\rowcolor{f2ecde}
\multicolumn{4}{l}{\emph{Cross-Modal Entity Binding}} \\
w/o Q-Former & 49.21\% & 51.63\% & 63.11\% \\
Vanilla Q-Former &  49.95\% & 52.75\% & 64.20\% \\
Entity-aware Q-Former   & 52.43\% & 56.31\% & 73.69\% \\
\rowcolor[gray]{0.95} Entity-aware Q-Former + ERA  & 53.17\% & 59.46\% & 73.85\% \\
\midrule
\rowcolor{f2ecde}
\multicolumn{4}{l}{\emph{Narrative-Aware Subject Arrangement}} \\
Direct Condition w/o MLLM & 51.07\% & 54.55\% & 70.14\% \\
\rowcolor[gray]{0.95} w/ MLLM arrangement & 53.17\% & 59.46\% & 73.85\%\\
\bottomrule
\end{tabular}
\label{tab:ablation}
\end{table}

\begin{figure}[!t]
\centering
\includegraphics[width=0.9\linewidth]{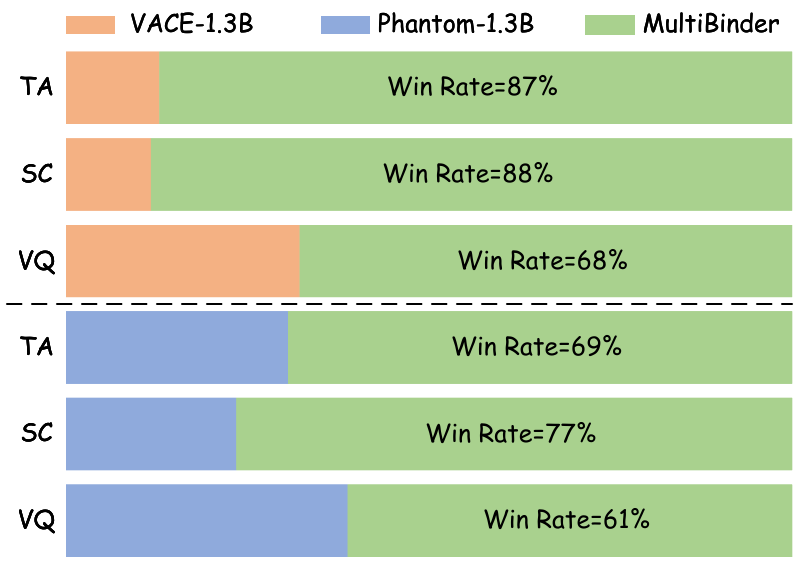}
\vspace{-0.09in}
\caption{\small User study between MultiBinder and other two state-of-the-art models across three dimensions. MultiBinder exhibits competent performance in subject consistency and text alignment.}
\label{fig:user_study}
\end{figure}

\subsection{User Study}
We further conduct a user study to compare MultiBinder against two state-of-the-art models built on the same backbone Wan2.1-T2V-1.3B: VACE-1.3B and Phantom-1.3B. The study evaluates human preference across three key aspects: text alignment (TA), subject consistency (SC), and visual quality (VQ). We recruited 30 participants from diverse backgrounds. Each participant is instructed to complete a questionnaire containing 50 pairwise comparison questions, selecting the preferred video from the two candidates for each aspect. As shown in \cref{fig:user_study}, MultiBinder achieves a higher win rate in all three dimensions against both baselines. Most notably, it attains the highest win rate in subject consistency dimension—88.3\% against VACE-1.3B and 76.5\% against Phantom-1.3B—confirming its superior ability to preserve subject identity in the generated videos.
The human preference results directly corroborate our quantitative findings and reinforce the superior efficacy of our ``bind-then-narrate'' paradigm in resolving the identity leakage issues that plague existing baselines. Furthermore, the high win rates in TA and VQ underscore that our structural constraints for identity preservation do not compromise but rather enhance the overall narrative coherence and visual fidelity of the generated content.

\section{Conclusion}
We present MultiBinder, a novel framework that introduces a ``bind-then-narrate'' paradigm to address the critical challenge of identity leakage and semantic entanglement in multi-subject personalized video generation. By explicitly decoupling entity binding from narrative arrangement prior to diffusion, our approach ensures that each custom subject retains its distinctive visual identity while correctly participating in a coherent dynamic scene. The core of our contribution lies in its structured two-stage conditioning design. The Cross-modal Entity Binding stage, empowered by an entity-aware Q-Former and refinement attention, establishes disentangled, phrase-aligned identity tokens. The subsequent Narrative-Aware Subject Arrangement stage leverages an MLLM as a scene director to synthesize contextualized ``who-is-doing-what'' features, which are then precisely injected into the video diffusion model. This design not only solves the targeted scalability problem but also yields general improvements in motion dynamics and overall naturalness. MultiBinder opens promising avenues for applications requiring faithful multi-character animation, such as personalized storytelling, interactive content creation, and dynamic advertising.

\begin{acks}
This work was supported in part by the Key Science \& Technology Project of Anhui Province No. 202523o09050002, the National Natural Science Foundation of China under grants 62472399 and Open Fund of APKL of BIIP, IAI, Hefei Comprehensive National Science Center under grants 24YGXT003.
\end{acks}

\bibliographystyle{ACM-Reference-Format}
\balance
\bibliography{mm2026}

\end{document}